\documentclass[letterpaper, 10 pt, conference]{ieeeconf}  

\IEEEoverridecommandlockouts                              

\usepackage[table]{xcolor}
\usepackage{pdfpages}
\usepackage{array}
\usepackage{tabu}
\usepackage{hhline}
\usepackage{amsmath}
\usepackage{url}
\usepackage{arydshln}
\usepackage{boldline}
\usepackage{hyperref}

\begin{document}




\title{V2CE: Video to Continuous Events Simulator}


\author{Zhongyang Zhang$^{1}$, %
Shuyang Cui$^{1}$, %
Kaidong Chai$^{2}$, %
Haowen Yu$^{4}$,%
\\Subhasis Dasgupta$^{1}$, %
Upal Mahbub$^{3}$, %
Tauhidur Rahman$^{1}$%
\thanks{$^{1}$University of California San Diego, San Diego, California, USA}
\thanks{$^{2}$University of Massachusetts Amherst, Amherst, Massachusetts, USA}
\thanks{$^{3}$Qualcomm Technologies, Inc., San Diego, California, USA}
\thanks{$^{4}$Independent Researcher, San Jose, California, USA}
}









\maketitle
\thispagestyle{empty}
\pagestyle{empty}

\begin{abstract}


Dynamic Vision Sensor (DVS)-based solutions have recently garnered significant interest across various computer vision tasks, offering notable benefits in terms of dynamic range, temporal resolution, and inference speed. However, as a relatively nascent vision sensor compared to Active Pixel Sensor (APS) devices such as RGB cameras, DVS suffers from a dearth of ample labeled datasets. Prior efforts to convert APS data into events often grapple with issues such as a considerable domain shift from real events, the absence of quantified validation, and layering problems within the time axis. In this paper, we present a novel method for video-to-events stream conversion from multiple perspectives, considering the specific characteristics of DVS. A series of carefully designed losses helps enhance the quality of generated event voxels significantly. We also propose a novel local dynamic-aware timestamp inference strategy to accurately recover event timestamps from event voxels in a continuous fashion and eliminate the temporal layering problem. Results from rigorous validation through quantified metrics at all stages of the pipeline establish our method unquestionably as the current state-of-the-art (SOTA). The code can be found at \href{https://bit.ly/v2ce}{bit.ly/v2ce}.

\end{abstract}


\maketitle

\section{Introduction}\label{Intro}

Neuromorphic cameras, also referred to as Dynamic Vision Sensors (DVS) or event cameras, have recently emerged as a significant area of interest in the field of robotics \cite{sandamirskaya2022neuromorphic, zhu2017event, li2021tracking} and computer vision \cite{gehrig2023recurrent, chamorro2023event, Baby_2017}. The exceptionally high optical event capture rate, high dynamic range, low yet adaptive power consumption, sparse output, and a dynamic vision scheme akin to mammalian perception contribute to their success in various computer vision applications \cite{Gehrig2021DSECAS,Rebecq2019HighSA,Mahlknecht2022ExploringEC}. These include feature tracking \cite{Seok2020RobustFT,Dong2021StandardAE}, optical flow estimation \cite{Pan2020SingleIO,Bardow2016SimultaneousOF}, as well as gesture and human pose estimation \cite{ZHANG2023126388,9959313}. Characteristically, DVS-based approaches tend to offer superior temporal resolution and quicker inference speeds.

Nevertheless, when compared with the Active Pixel Sensor (APS, standard RGB camera), the DVS emerges as a relatively novel vision sensor. Also, compared to APS frames, labeling DVS data is quite challenging, as events captured are sparse, and inactivative objects trigger few events. Consequently, there is a scarcity of large-scale annotated DVS datasets, which are significantly harder to procure than APS data. Unlike APS data that can be readily generated by mobile devices and obtained from the internet, DVS data collection necessitates specific hardware such as a DVS camera and a laptop. Moreover, dataset collection typically proves to be time-consuming and expensive, thereby posing significant challenges to the acquisition of large-scale DVS datasets. Finally, it is neither practical nor cost-effective to recreate every existing APS dataset for DVS. In this work, we present an optimized video-to-event conversion algorithm that can effectively mimic the nonlinear characteristics of a DVS camera with high fidelity.

\begin{figure}
    \centering
    \includegraphics[width=\linewidth]{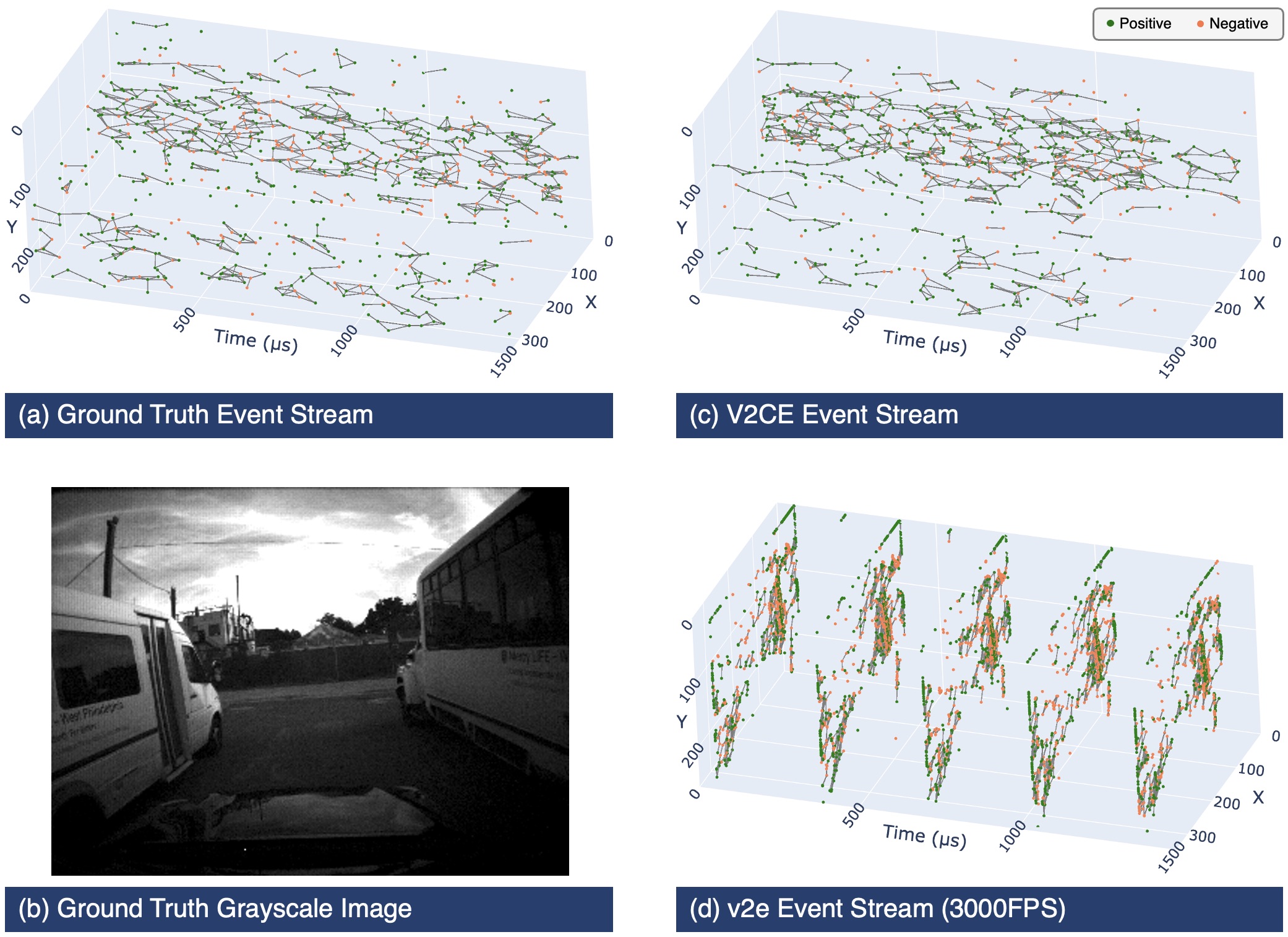}
    \caption{K-nearest neighbor graph comparison from events in $(x,y,t)$ space. The v2e event stream is generated with input video upsampled to 3000FPS.}
    \label{fig:es-vis}
\end{figure}

There are a few existing works trying to bridge the gap between the APS frames and events, like ESIM \cite{pmlr-v87-rebecq18a, Gehrig_2020_CVPR}, v2e \cite{Hu2021-v2e-cvpr-workshop-eventvision2021}, and EventGAN \cite{zhu2019eventgan}. These methods can be roughly divided into two genres: rule-based \cite{Hu2021-v2e-cvpr-workshop-eventvision2021, pmlr-v87-rebecq18a, Gehrig_2020_CVPR} and learning-based \cite{zhu2019eventgan}. However, the former abandoned recovering the lost information due to the dynamic range gap between standard APS and DVS \cite{10168206, liu2022lowlight}, while the latter doesn't consider the characteristics difference between these two types of cameras. Lastly, none of these previous works have ever discussed the last mile problem: how to convert the generated event voxels or the events number into realistic and accurate raw event streams. All these works directly apply either random or even sampling, which is clearly suboptimal.

\begin{figure*}[ht]
    \centering
    \includegraphics[width=\linewidth]{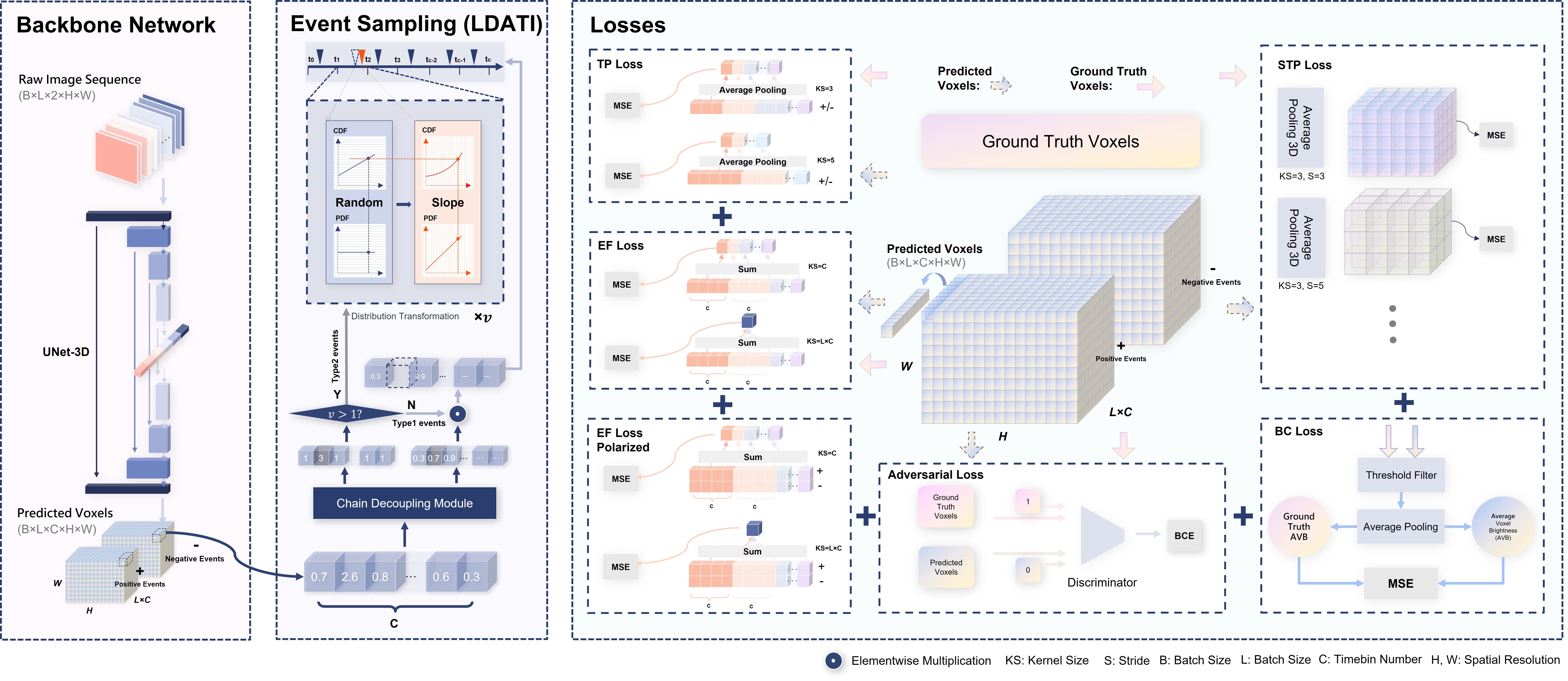}
    \vspace{-0.5cm}
    \caption{Proposed Motion-Aware Event Voxel Prediction Pipeline and Hybrid Loss Structure: Our method consists of two main stages. The Backbone 3D UNet encodes input frame pair sequences and generates event frames. The Event Sampling Module, subdivided into chain decoupling and distribution transformation modules, calculates event counts and in-voxel time, then redistributes events in Type2 voxels. The loss functions for training, displayed on the right, include STP, TP, ADV, BC, and EF Losses, which are elaborated in Section \ref{sec:stage1}. }
    \label{fig:overall-pipeline}
\end{figure*}

Notably, while these methodologies successfully convert videos to events, the resulting events continue to reside in a series of discrete temporal layers. Fig. \ref{fig:es-vis} illustrates a 3D visualization for ground truth events and events generated by V2CE and v2e \cite{Hu2021-v2e-cvpr-workshop-eventvision2021}. For v2e, It's apparent that all the generated events share a series of discrete timestamps, instead of spreading across the time axis in a continuous fashion like real DVS recordings. This discrepancy is often negligible when temporal accumulation-based methods are utilized in subsequent task preprocessing, as the temporal information is collapsed anyway. However, for tasks that are sensitive to timestamps distribution, such as Graph Neural Network (GNN) \cite{Scarselli2009TheGN, schaefer2022aegnn, 10166491, slide_gcn} and Spiking Neural Network (SNN) \cite{Tavanaei2018DeepLI, Deng2022TemporalET, cordone2021learning, zhu2022eventbased}, this issue could prohibit using generated synthetic data as pretraining dataset, since these data has a significant domain shift compared to real events.

In this paper, we introduce a new Video-to-Continuous Events (V2CE) framework that tackles video-to-event conversion challenges via a two-stage pipeline. Our main contributions include:

\begin{enumerate}
\item We introduce a specialized suite of loss functions tailored for the video-to-event voxel task, thereby achieving SOTA performance.
\item We develop a novel statistics-based local dynamics-aware timestamp inference algorithm that enables the smooth transition from event voxels to event streams, outperforming existing baseline methods.
\item We establish the first set of metrics grounded in DVS event characteristics, allowing for robust quantitative evaluation in both the video-to-event voxel and the voxel-to-event stream phases.
\item Through rigorous evaluation against established baselines, we demonstrate that V2CE significantly outperforms them across all metrics, and the simulated events' count strictly matches the ground truth. Our comprehensive results analysis further underscores that V2CE is not only the SOTA but also the first option for generating continuous event streams.
\end{enumerate}





\section{Proposed System}\label{pipeline}


The proposed Video to Continuous Events (V2CE) simulator pipeline (outlined in Fig. \ref{fig:overall-pipeline}) consists of two stages: motion-aware event voxels prediction, and voxels to continuous events sampling.

\subsection{Stage1: Motion-Aware Event Voxel Prediction}\label{sec:stage1}

The Motion-Aware Event Voxel prediction aims to transform the APS video into a 3D voxel grid where the video data is temporally upsampled. The temporal resolution is increased by a significant margin and the event sequence is represented in a spatio-temporal $xyt$ coordinate system. We use a simple UNet-based architecture to design this module. 

However, the main challenge of this task is to preserve the temporal continuity and the microstructure compatibility of event voxels. High-fidelity event voxel reconstruction requires information about nonlinear dynamics of light intensity changes and object movements (e.g., acceleration or higher order moment). While any linear assumption invariably leads to suboptimal video-to-event conversion performance, all prior work in this space only used an adjacent frame pair to infer the events between them. Since no hint is available to infer the nonlinear dynamics, all baseline methods were essentially doing linear interpolation between the input APS frame pair.

Therefore, we advocate using longer frame sequence instead of frame pairs to serve as the input of the model, and help local temporal information flow properly during the inference. Bear these considerations in mind, we modified a 3D UNet model and use a sequence of 16 frame pairs as the input to the model.


Further complicating the task, event and APS cameras differ in dynamic ranges, which affects information compression in overexposed and underexposed areas. Additionally, both camera types have adjustable parameters such as exposure, ISO, and aperture, which can be dynamically tuned to adapt to varying environments. This renders the video-to-event voxel prediction a time-varying task, making a straightforward one-to-one mapping between APS video frames and event voxels especially challenging.

To address this complex task, a key contribution of our work is the development of a hybrid loss function composed of five distinct losses, which we briefly describe below.

Denote the input to the model as $I\in \mathcal{R}^{(B, L, 2, H, W)}$, where the five dimensions represent the batch size, sequence length, and spatial resolution. Then the output event voxels satisfy $V\in \mathcal{R}^{(B,L,2\times C, H, W)}$, where $C$ represents the timebin number between two frames, and the third dimension has a shape of $2\times C$ since events of different polarities are also separated. All losses take ground truth voxels and predicted voxels as input.

The first loss to introduce is the \textbf{Spatial-Temporal-Pyramid Loss} (\textbf{STP Loss}, $\mathcal{L}_{STP}$). STP loss takes the entire concatenated voxel with a shape of $(B, L\times C, H, W)$ and applies a series of 3D Average Poolings with varying kernel sizes and strides. This produces more compact representations of both ground truth and predicted event voxels. The STP Loss encourages the model to extract multi-scale information from adjacent voxels, enhancing its robustness against noise by applying coarse supra-voxel matching. Formally, the STP Loss is defined as:
\begin{equation}
\mathcal{L}_{STP} = \sum_{k\in \mathcal{K}, s \in \mathcal{S}} w_{k,s} \cdot \left\| \mathcal{P}^{3D}_{k,s}(V_{GT}) - \mathcal{P}^{3D}_{k,s}(V_{pred}) \right\|_2^2
\end{equation}
Where $\mathcal{P}^{3D}_{k,s}(V)$ denotes 3D average pooling operation applied to voxel $v$ with a kernel size $k$ and stride $s$, $\mathcal{K}$ represents the set of all kernel sizes used in the pooling operations, $\mathcal{S}$ represents the set of all strides used in the pooling operations, and $w_{k,s}$ denotes the weights for each combination of kernel size $k$ and stride $s$.




\textbf{Temporal-Pyramid Loss} (\textbf{TP Loss}, $\mathcal{L}_{TP}$) is designed to prioritize neighboring events, which are crucial for voxel-level event reconstruction. We employ 1D average pooling along the time axis using varying kernel sizes and strides on both ground truth and predicted event voxels, followed by an L2 loss calculation. Formally, the TP Loss is defined similarly to STP Loss:
\begin{equation}
\mathcal{L}_{STP} = \sum_{k,s \in \mathcal{K,S}} w_{k,s} \cdot \left\| \mathcal{P}^{T}_{k,s}(V_{GT}) - \mathcal{P}^{T}_{k,s}(V_{pred}) \right\|_2^2
\end{equation}
Where $\mathcal{P}^{T}_{k,s}$ denotes 1D average pooling along the time axis.

Similarly, \textbf{Event Frame Loss} (\textbf{EF Loss}, $\mathcal{L}_{EF}$) compresses the time axis by summing timebins between adjacent frames or across the entire frame sequence along the time. This addresses the issue of sparsity in voxels and ensures better and aligned information flow between generated event frames and the input frame sequence. Event frames with and without polarity consideration are separately counted in the loss calculation.
\begin{equation}
\begin{aligned}
\mathcal{L}_{EF}= \left\| \mathcal{S}_{C}(V_{GT}) - \mathcal{S}_{C}(V_{pred}) \right\|_2^2+\\
\left\| \mathcal{S}_{LC}(V_{GT}) - \mathcal{S}_{LC}(V_{pred}) \right\|_2^2
\end{aligned}
\end{equation}
Where \( \mathcal{S}_{C}(\cdot) \) and \( \mathcal{S}_{LC}(\cdot) \) denotes the compression operation that sums over timebins \( C \) between adjacent frames and the entire frame sequence \( LC \) respectively.

\textbf{Adversarial Loss} (\textbf{ADV Loss}, \(\mathcal{L}_{ADV}\)) aims to enhance the realness of our generated event voxels. Utilizing both ground truth and predicted voxels as real and fake samples respectively, the discriminator is trained for optimal distinction. To prevent \(\mathcal{L}_{ADV}\) from becoming unbounded, the generated event voxels strive for high similarity with real voxels to effectively deceive the discriminator.


As previously noted, the relationship between APS frames' pixel brightness and the event number between frame pairs is not static, necessitating a dynamic, semantics-based modeling of intrinsic camera parameters. This complexity arises because APS captures brightness as \(\phi(I)\), while DVS records \( \log(\phi(I)) \), where \(I\) is the scene's absolute brightness and \(\phi(I)\) represents the effect of camera parameters. Given that multiple intrinsic parameters affect \(\phi\), a fixed linear mapping is untenable. To address this, we introduce \textbf{Brightness-Compensation Loss} (\textbf{BC Loss}, \(\mathcal{L}_{BC}\)), which compute the average brightness \(I_a\) of voxels exceeding a threshold \(\beta\), and align this \(I_a\) with that of the ground truth voxels. we define the average brightness \(I_a\) as:
\begin{equation}
I_a(V) = \frac{\sum_{v \in V, v > \beta} v}{|\{v \in V : v > \beta\}|}
\end{equation}
Where \(\beta\) serves as a threshold to consider voxels that exceed a certain brightness. Given this, the BC Loss between ground truth voxels \(V_{GT}\) and predicted voxels \(V_{pred}\) is:
\begin{equation}
\mathcal{L}_{BC} = \left\| I_a(V_{GT}) - I_a(V_{pred}) \right\|_2^2
\end{equation}
At last, all these losses are combined together with a separate weight factor $\alpha$ (which is learned by grid search). The complete loss formula is:
\begin{equation}
\begin{aligned}
    \mathcal{L} = &\alpha_{STP}\mathcal{L}_{STP} + \alpha_{TP}\mathcal{L}_{TP} + \\
    &\alpha_{EF}\mathcal{L}_{EF} + \alpha_{ADV}\mathcal{L}_{ADV} + \alpha_{BC}\mathcal{L}_{BC}
\end{aligned}
\end{equation}
Once trained, the motion-aware event voxel prediction generates 10 voxels per pixel between consecutive grayscale video frames at 30 fps. This effectively maps each frame with dimensions $H \times W$ to an event voxel grid of dimensions $(2 \times 10) \times H \times W$. Upon evaluation with the entire MVSEC dataset \cite{zhu2018multivehicle}, we found this ten-fold upscaling to be adequate. Specifically, only approximately 2.30\% of the voxels are non-zero, and among these, a mere 6.66\% exceed one. This validates that partitioning the time range between two frames into 10 timebins is sufficient for capturing the event stream.

\begin{figure}[h]
    \centering
    \includegraphics[width=\linewidth]{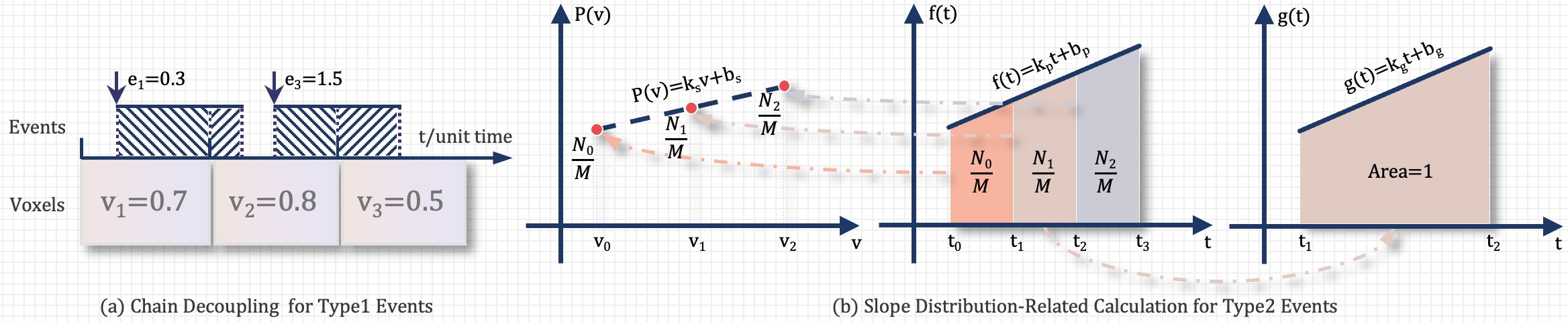}
    \vspace{-0.5cm}
    \caption{Visualization for two stages in LDATI.}
    \label{fig:stage2-math}
\end{figure}

\subsection{Stage2: Voxels to Continuous Events Sampling}\label{sec:stage2}


This second task recovers precise event timestamps from Stage1's event voxels. Previous research used basic sampling methods, such as random or uniform sampling. We introduce \textbf{Local Dynamics-Aware Timestamp Inference (LDATI)}, an advanced technique considering each voxel's event firing trends. It achieves a significantly lower error metric, which is only \textit{\textbf{3.5\%}} of the best traditional method.

Event voxels aim to discretize temporally contiguous events into a dense tensor, suitable for deep learning inference. Rather than merely counting the event numbers in each voxel (with the temporal resolution of $\delta$), the generation of event voxels also preserves the relative temporal information of events within the timebin. Each event influences the voxel series for a short and finite duration which can be characterized by a continuous-time unit step signal (with an on-time duration same as $\delta$). The value of each voxel is determined by integrating all the step signals for all events within a voxel's designated time range. The sum of all voxels at the same pixel location equals the total number of events occurring within that time frame. This allows the voxel to summarize the total number of events and their relative times with a single number.

Considering the inverse process of event voxel generation, let \( v \) be the value of a voxel and \( v' \) be its value after removing the influences of events from the preceding voxel. If only one event is fired in the current voxel, its relative occurrence time within the voxel can be determined by \( e = \lceil v' \rceil - v' \). This event will then exert an \( e \)-unit influence on subsequent voxels. Starting from the first voxel, where \( v = v' \), we can iteratively deduce the event count and their relative positions in each voxel, a process we term as Chain Decoupling (Fig. \ref{fig:stage2-math}.a). This computation is deterministic, as it merely reconstructs the inherent temporal information in the voxels.

Voxels are Type1 if \( v' \leq 1 \) and Type2 if \( v' > 1 \). As per Section \ref{sec:stage1}, event voxels are sparse. Type1 voxels use Chain Decoupling; Type2 requires alternatives. Motion and changes happen in a continuous manner and don't change abruptly under natural condition. Thus, if the preceding voxel has a greater value than the succeeding one, events in the intervening voxel are more likely to be biased towards the former, and vice versa.



To accurately model this phenomenon, we assume that each voxel and its neighboring voxels conform to a slope distribution described by the Probability Density Function (PDF) \(f(t) = k_pt+b_p\). Given that the event timestamp distribution within a voxel is primarily influenced by its temporal neighbors, a simpler slope distribution suffices for both accuracy and computational efficiency. This PDF allows us to estimate event timestamps while accounting for local dynamics, outperforming random sampling approaches.

Let the voxel value in three adjacent voxels be denoted as \(N_0\), \(N_1\), and \(N_2\), and their sum as \(M\). Our objective is to derive the expression for their PDF \(f(t)\) using these known variables. If \(g(t)\) represents the PDF of the event timestamp distribution conditional on \(t\) being in the central voxel \(v_1\), we have:
\begin{align}\label{eq:gt}
    g(t) = f_{T|V}(t|v) = \frac{f_{T,V}(t, v)}{P(V=v)} = \frac{f(t)}{P(V=v)}
\end{align}
In this formulation, \(T\) and \(V\) denote the exact event timestamp and its corresponding timebin, respectively. The relationship among \(f(t)\), \(g(t)\), and \(P(v)\) is visually explained in Fig. \ref{fig:stage2-math}.b. It can be readily shown that \(P(v)\) also follows a linear formula with a slope \(k_s = \delta k_p = \delta k_g\). Given \(k_s = \frac{N_2 - N_0}{2\delta M}\), the expression for \(g(t)\) can be derived accordingly.
\begin{align}
    g(t)=k_pt+1/\delta-\delta k_p/2
\end{align}
where \(k_p = \frac{N_2 - N_0}{2\delta^2 M}\). While \(g(t)\)-based sampling could be applied individually on each voxel, this would be computationally expensive due to varying distribution formulas and event numbers. However, we can optimize this process through distribution transformation. By initially sampling from a uniform distribution with PDF \(\gamma(u)=1/\delta\), and then converting it to the desired slope distribution via matrix operations, we significantly accelerate the sampling process. This conversion can be achieved using the inverse cumulative distribution function (CDF) method as follows:
\begin{align}
    t= (-b_p +\sqrt{b_p^2+2k_pu})/k_p
\end{align}

\section{Experiments and Results}\label{results}
\subsection{Stage1: Motion-Aware Event Voxel Prediction}

\begin{table*}[ht]
\centering
\caption{The quantitative comparison between our method and all baseline methods on the testing set.}
\begin{tabular}{lllllllll}
\hlineB{2}
\hlineB{2}
\cellcolor{gray!20}\textbf{Method}   & \cellcolor{gray!40}\textbf{TPF1$\uparrow$}   & \cellcolor{gray!40}\textbf{TF1$\uparrow$} & \cellcolor{gray!20}\textbf{RF1$\uparrow$} & \cellcolor{gray!20}\textbf{TPAcc$\uparrow$} & \cellcolor{gray!20}\textbf{TAcc$\uparrow$} & \cellcolor{gray!20}\textbf{RAcc$\uparrow$} & \cellcolor{gray!20}\textbf{PMSE-2$\downarrow$} & \cellcolor{gray!20}\textbf{PMSE-4$\downarrow$} \\
ESIM \cite{Gehrig_2020_CVPR}     & 0.3293 & 0.2670 & 0.1301 & 0.6240 & 0.7200 & 0.8801 & 3.03512 & 1.54268   \\
EventGAN\cite{zhu2019eventgan} & 0.2398 & 0.1544 & 0.0520 & 0.1773 & 0.1773 & 0.1530 & 0.05779 & 0.03030   \\
V2E\cite{Hu2021-v2e-cvpr-workshop-eventvision2021}      & 0.3656 & 0.3197 & 0.1604 & 0.7214 & 0.8207 & 0.9269 & 0.04929 & 0.02618   \\
Ours     & \textbf{0.5323} & \textbf{0.5058} & \textbf{0.3014} & \textbf{0.8891} & \textbf{0.9266} & \textbf{0.9709} & \textbf{0.00267} & \textbf{0.00074}    \\ \hlineB{2} \hlineB{2}
\end{tabular}
\label{tab:baseline}
\end{table*}

\begin{figure}
    \centering
    \includegraphics[width=\linewidth]{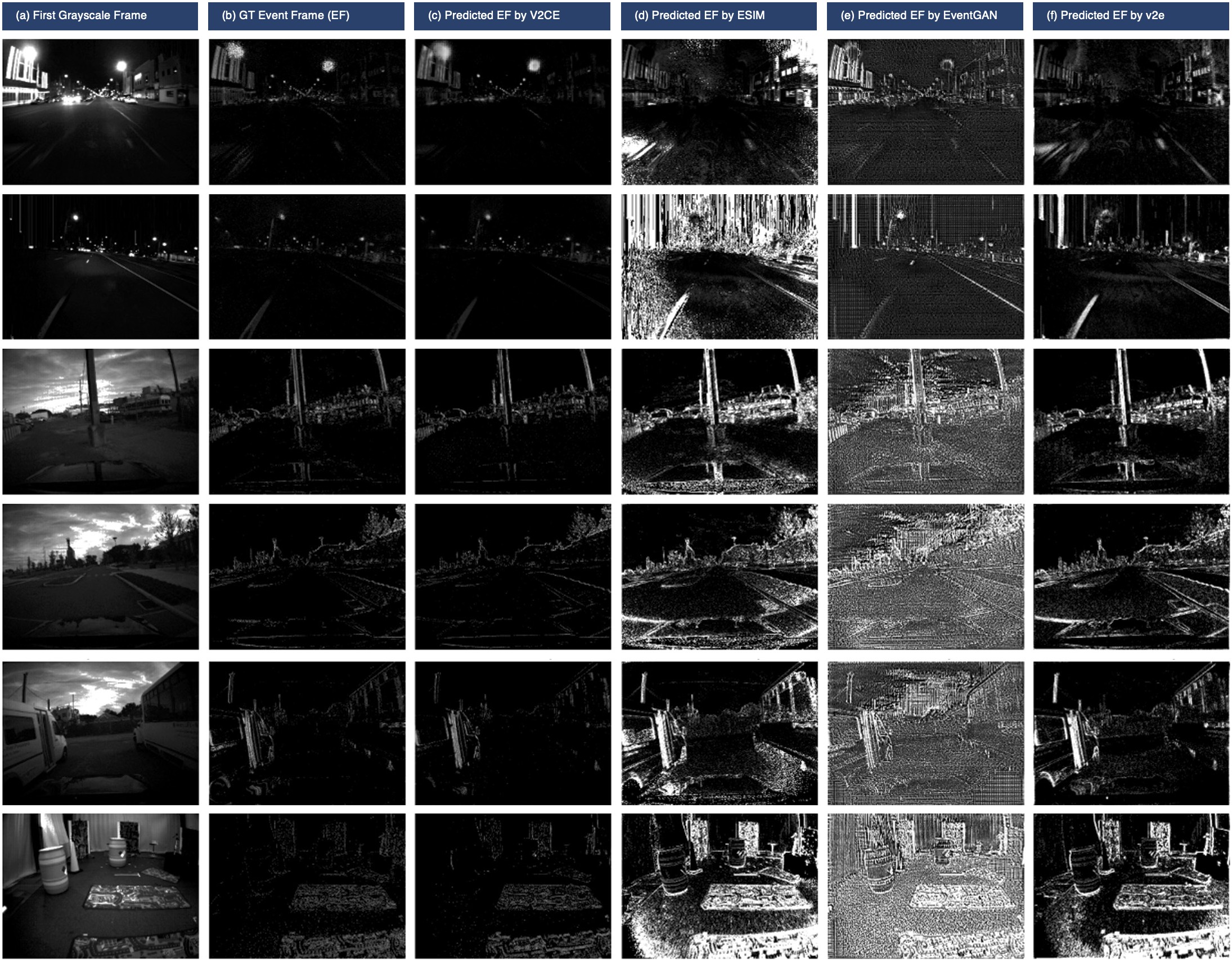}
    \caption{Event frame comparison between V2CE and baseline methods. Event frames
are clipped to the maximum value of their corresponding ground truth event
frames and then normalized.}
    \label{fig:2d-comp}
\end{figure}

Existing video to event simulation methodologies primarily rely on qualitative assessments and consequently lack in standardized objective quantitative evaluation metrics. To address these challenges, we introduce the following novel metrics tailored for this specific task:

1) \textbf{PMSE (Pooling MSE)} aims to compare the predicted voxel cube with the ground truth by employing a 3D average pooling (with kernel and stride of 2 or 4) and then estimating the Mean Square Error (MSE). The 3D average pooling extracts a slightly higher-level summary of the sparse voxel cube. 2) \textbf{TAcc} collapses the \textit{T}ime axis of the 3D voxel cube to convert to a 2D frame for each polarity, apply a thresholding (with a value of 0.001) to convert it to two binary 2D frames for each polarity, and lastly estimate the accuracy by performing the binary matching. 3) \textbf{TPAcc} is similar to TAcc, except it further collapses the event \textit{P}olarity by accumulating the voxel cubes of two polarities. 4) If no temporal or polarity-wise accumulation is performed, the metric is termed \textbf{RAcc} (Raw Accuracy). 5) \textbf{TPF1, TF1, RF1}: we follow the same procedure as TAcc, TPAcc, or RAcc except F1 scores are calculated instead. Given that event voxels are generally sparse, the F1 score is more representative. TAcc, TPAcc, TF1, and TPF1 aim to evaluate high-frequency spatial patterns, while RAcc and RF1 provide a straightforward way to check the absolute voxel occupancy condition. TPF1 and TF1 are selected as the main metrics.

\subsubsection{Comparison with Existing Works}



We assess our video-to-events model against baseline methods using metrics that encompass both temporal and spatial aspects, as previously outlined. Consistent testing, training, and validation sets are employed for all methods. For EventGAN \cite{zhu2019eventgan}, we set its timebin number to 10 as well during retraining to align with our framework, while maintaining other settings. ESIM \cite{Gehrig_2020_CVPR} and V2E \cite{Hu2021-v2e-cvpr-workshop-eventvision2021}, as non-model-based approaches, were executed with default configurations. As shown in Table \ref{tab:baseline}, our model excels across all metrics and achieve SOTA.

\begin{figure}[]
    \centering
    \includegraphics[width=\linewidth]{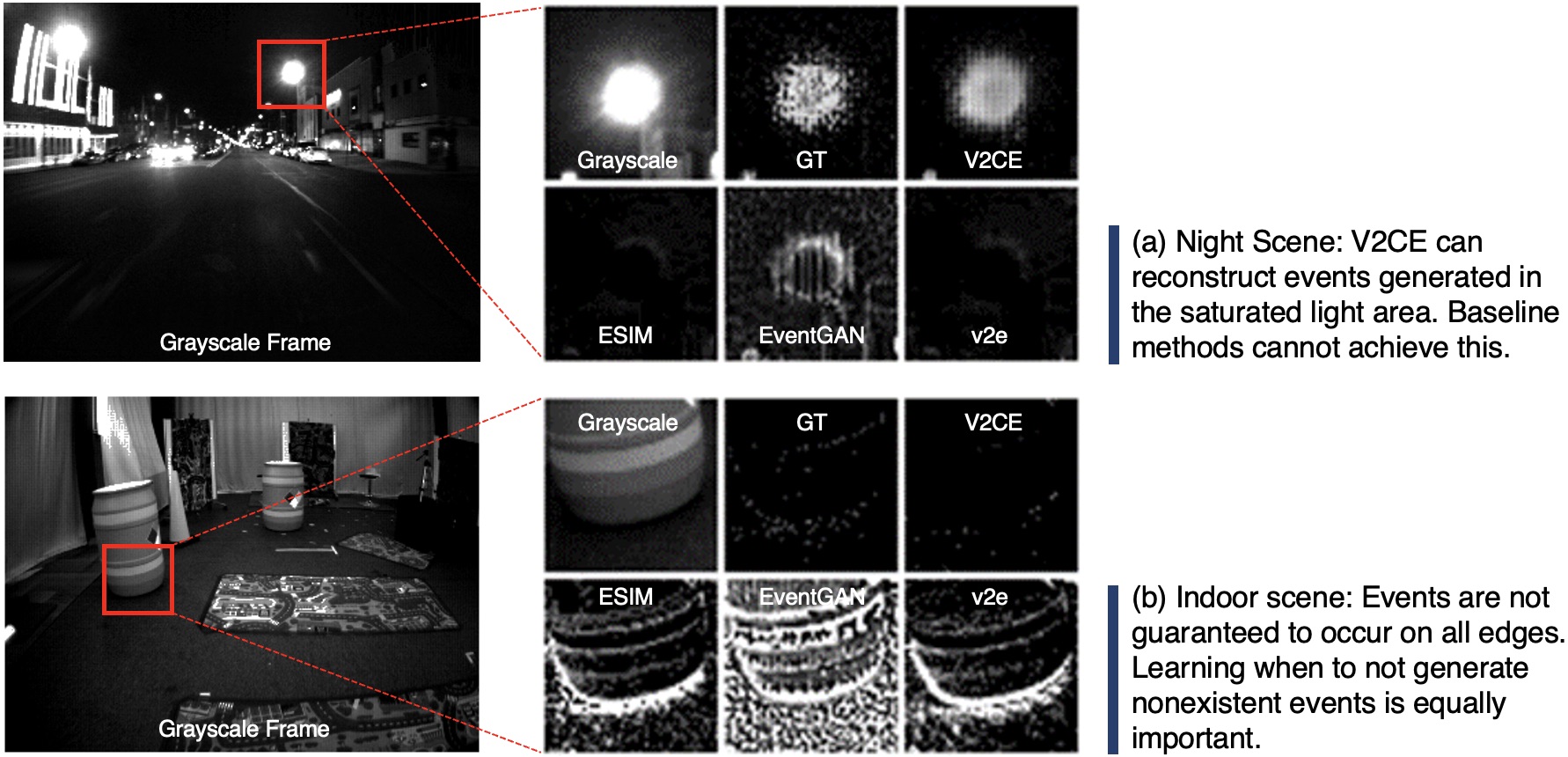}
    \caption{Zoom-in comparison on event frame details.}
    \label{fig:zoomin}
\end{figure}

\begin{figure}
    \centering
    \includegraphics[width=\linewidth]{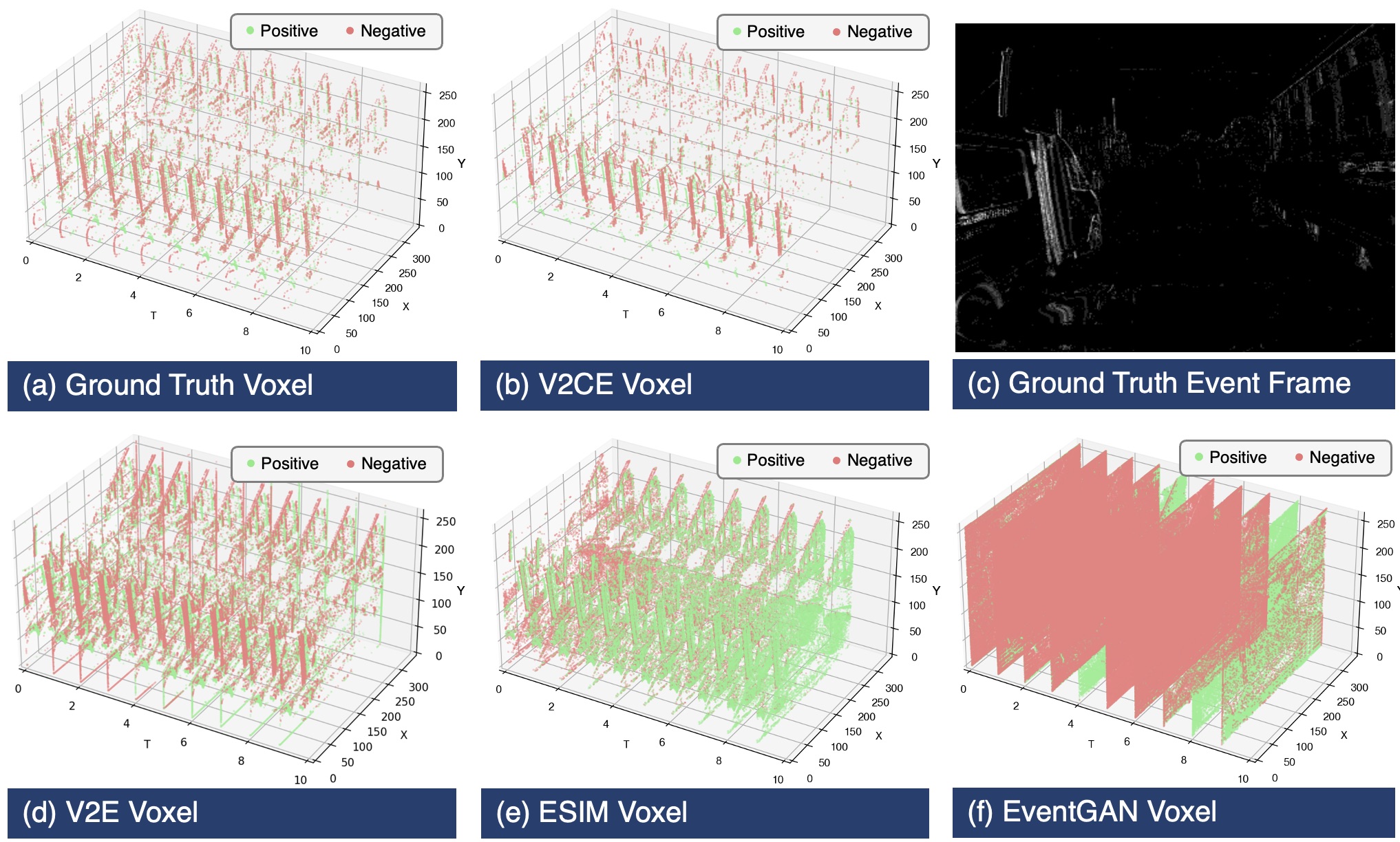}
    \caption{Voxel Cube comparison. (Notice: Voxels, not raw event stream.)}
    \label{fig:voxel_3d}
\end{figure}

Qualitative comparison can be found in Fig. \ref{fig:2d-comp}, \ref{fig:zoomin}, and \ref{fig:voxel_3d}. Fig. \ref{fig:2d-comp} shows a 2D event frame-level comparison between all the video-to-event voxel methods, and all event frames are clipped to the maximum value of the ground truth event frame and then normalized. Thanks to our specially designed losses, our predicted event frames are loyal to the actual brightness level, which means the events' number generated is close to the actual events' number. However, all the baseline methods tend to generate much more events in general, and the details don't match well. Due to the unstable and noise-prone nature of the underexposure part of the grayscale images, baseline methods generally behave poorly in night scenes. For the overexposed region of a scene, since the pixel value has saturated and the detailed information in these regions is lost. Without a good semantic understanding, this lost information can never be recovered. Moreover, all baseline methods behave quite aggressively when it comes to any edges, but different from a simple edge detector, a video-to-event converter not only needs to learn when to generate events but also when not to. A zoom-in comparison can be found in Fig. \ref{fig:zoomin}.

To better visualize the performance of different methods on the time axis, we also generated a 3D plot in the $xyt$ space, as shown in \ref{fig:voxel_3d}. If the value of a voxel is greater than 0.001, one node will be placed in the corresponding location. It's clear that our prediction also matches the ground truth voxels significantly better both spatially and temporally.


\begin{table}[bp]
\centering
\caption{The ablation study results for all our proposed losses.}
\begin{tabular}{llllll}
\hlineB{2}
\hlineB{2}
\cellcolor{gray!20}\textbf{Ablation Item} & \cellcolor{gray!40}\textbf{TPF1} & \cellcolor{gray!40}\textbf{TF1} & \cellcolor{gray!20}\textbf{RF1} & \cellcolor{gray!20}\textbf{PMSE-2} & \cellcolor{gray!20}\textbf{PMSE-4} \\
STP Loss      & 0.5253 & 0.4999 & 0.3041 & 0.00279 & 0.00078 \\
TP Loss       & 0.5170 & 0.4794 & 0.2528 & 0.00314 & 0.00084 \\
EF Loss       & 0.4638 & 0.4461 & 0.2938 & 0.00248 & 0.00075 \\
ADV Loss      & 0.3583 & 0.2817 & 0.1662 & \textbf{0.00225} & \textbf{0.00068} \\
BC Loss       & 0.5253 & 0.4983 & \textbf{0.3070} & 0.00277 & 0.00076 \\
w/o Ablation    & \textbf{0.5323} & \textbf{0.5058} & 0.3014 & 0.00267 & 0.00074 \\ \hlineB{2}
\hlineB{2}
\end{tabular}
\label{tab:ablation}
\end{table}

\subsubsection{Ablation Study}

As indicated in Table \ref{tab:ablation}, the removal of any loss functions leads to a decline in our primary metrics, TPF1 and TF1. Eliminating $\mathcal{L_{STP}}$ or $\mathcal{L_{BC}}$, which focuses on high-level matching and compensation, slightly improves the voxel-wise metric RF1 but negatively impacts all other metrics. Deactivating the $\mathcal{L_{ADV}}$ results in a significant drop in all F1 metrics, despite a concurrent decrease in PMSE losses. This suggests that while the absence of $\mathcal{L_{ADV}}$ may improve voxel value matching, it significantly deviates the generated voxels from real voxels' distribution. Hence, each proposed loss function positively influences the final performance.

\subsubsection{Training Details}

We trained our neural network using the Adam optimizer, setting the learning rate to 0.001 and running the training for 100 epochs on our dataset. This project is built on top of PyTorch Lighting. We employed the Multi Vehicle Stereo Event Camera (MVSEC) dataset \cite{zhu2018multivehicle}, which is one of the largest and most commonly used datasets for DVS-based research. The dataset was randomly partitioned into training, validation, and testing subsets, following an 80\%/10\%/10\% distribution. Each data entry in the dataset comprises 17 sequential grayscale frames, along with the associated 16 event packets in between.

The size of our Stage1 model is 52.9MB, and it requires 779.17 GFlops to perform inference on a single image pair sequence. Each sequence has a length of 16, which corresponds to 0.53 seconds of video length when the video's frame rate is 30Hz. When evaluated on an A10 graphics card, the model's average inference time for a 16-pairs sequence was 312.83 milliseconds, and the corresponding sampling time in Stage2 was 106.97ms. This allows us to generate events from approximately 39 frames per second, meeting the criteria for real-time performance.

\subsection{Stage2: Voxel to Continuous Event Sampling}

\begin{figure}
    \centering
    \includegraphics[width=\linewidth]{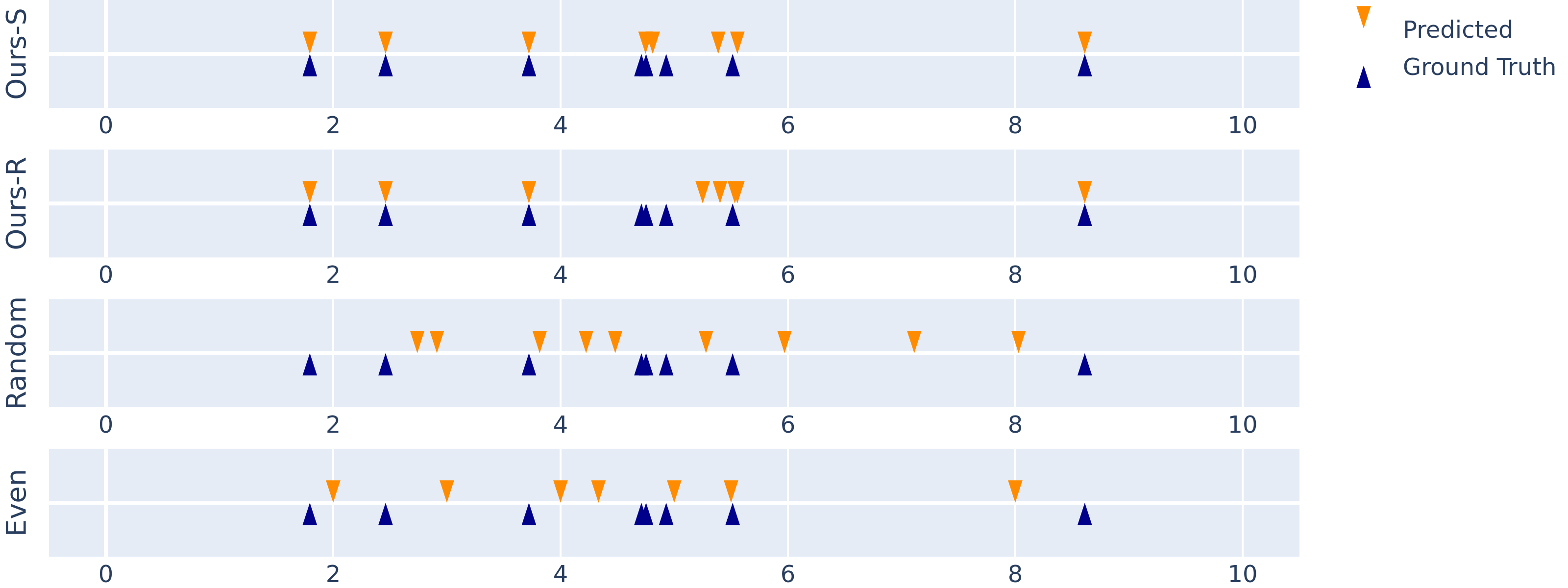}
    \caption{Qualitative comparison between all voxel-to-event sampling methods given a randomly generated event stream (blue upward wedges). In this figure, ``Ours-S'' and ``Ours-R'' represent chain decoupling with slope distribution and chain decoupling with random distribution for Type2 events.}
    \label{fig:stage2-vis}
\end{figure}

To evaluate event voxel-to-event stream generation, we introduce three metrics. \textbf{Mean Event Timestamp Error (METE)} quantifies the average temporal discrepancy, in microseconds, between each ground truth and the nearest predicted event at the same pixel. Ground truth events without a nearby counterpart within $3\delta$ on the time axis are termed ``overflow events,'' and their error is capped at $3\delta$. \textbf{Number of Overflowed Events (NOE)} quantifies the number of such extreme errors. To mitigate potential bias from oversampling, we introduce \textbf{Predicted/Ground Truth Event Number Ratio (PGER)}, aiming for a value as close to 1 as possible. By multiplying the PGER when it is greater than one, we can get the calibrated METE and NOE, which are called \textbf{C-METE} and \textbf{C-NOE}.


\begin{table}[ht]
\centering
\caption{Testing set event voxel-to-event stream results comparison.}
\begin{tabular}{lllll}
\hlineB{2}
\hlineB{2}
\cellcolor{gray!20}\textbf{Method}  & \cellcolor{gray!20}\textbf{Type2 Events}    &  \cellcolor{gray!20}\textbf{C-METE} & \cellcolor{gray!20}\textbf{C-NOE} & \cellcolor{gray!20}\textbf{PGER} \\
Even    & Even Sampling   &  4003.528       & 2362       & 1.000 \\
Random  & Random Sampling &  3657.325       & 2350       & 1.000 \\
LDATI & Random Sampling    &  851.679        & 638        & 1.000 \\
LDATI  & Slope Distribution  & \textbf{142.884}        & \textbf{0}                            & 1.000                     \\ \hlineB{2}
\hlineB{2}
\end{tabular}
\label{tab:stage2-result}
\end{table}



For baselines, we consider two standard techniques: random and even samplings. Let \( v \) be the input voxel value. In both methods, the estimated event number \( \tilde{v} \) in a voxel is computed as \( \tilde{v} = \lfloor v \rfloor + \text{Bernoulli}(v - \lfloor v \rfloor) \). In random sampling, \( \tilde{v} \) events are sampled randomly within the timebin. In even sampling, we find the maximum \( M = \max(\tilde{v}) \) across all voxels, divide the timebin into \( M \) equal sub-timebins, and sequentially place \( \tilde{v} \) events starting from the leftmost sub-timebin. A qualitative comparison is shown in Fig. \ref{fig:stage2-vis}.


To rigorously evaluate our LDATI sampling method, we conducted an experiment on ground truth event streams from our test set, converting them to event voxels as outlined in Section \ref{sec:stage2}. We then applied various sampling techniques for event reconstruction and computed the predefined metrics. The results, detailed in Table \ref{tab:stage2-result}, confirm a PGER metric of 1 for all methods, eliminating under- or over-sampling concerns. LDATI significantly outshines the baselines, with a C-METE metric just \textbf{3.5\%} that of the even sample and a NOE of zero throughout the test set. Substituting our slope distribution sampling with random sampling for Type2 events led to a \textbf{6}$\times$ higher C-METE error, underscoring the efficacy of slope distribution sampling.

In our study, we implemented a two-stage evaluation and refined the metrics. Although Stage 1 introduces notable voxel-level errors, skewing the overall error relative to ground truth, our V2CE pipeline demonstrates its efficacy as shown in Table \ref{tab:full-stage-result}. Unlike prior methods that tend to oversample events substantially (up to \textbf{53.3}$\times$ more), leading to significant divergence from the true event distribution, V2CE closely matches the ground truth in event count while simultaneously achieving optimal accuracy metrics.

\begin{table}[ht]
\centering
\caption{Testing set video-to-event stream results comparison.}
\begin{tabular}{llccc}
\hlineB{2}
\hlineB{2}
\cellcolor{gray!20}\textbf{Stage1}  & \cellcolor{gray!20}\textbf{Stage2}  & \cellcolor{gray!20}\textbf{C-METE} & \cellcolor{gray!20}\textbf{C-NOE} &  \cellcolor{gray!20}\textbf{PGER}\\
EventGAN\cite{zhu2019eventgan} & Random  & 395057.269 & 372383 & 45.280 \\
ESIM\cite{Gehrig_2020_CVPR} & Even  & 876386.946 & 1346561 & 54.299 \\
v2e\cite{Hu2021-v2e-cvpr-workshop-eventvision2021} & Even  & 63738.935 & 82238 & 5.145 \\
V2CE & Even  & 11131.586 & 12568 & 0.720 \\
V2CE & Random  & 11079.564 & 12487 & 0.720 \\
V2CE & LDATI & \textbf{10039.262} & \textbf{10364} & \textbf{0.997}\\ \hlineB{2} 
\hlineB{2}
\end{tabular}
\label{tab:full-stage-result}
\end{table}




\section{Conclusion}



In this paper, we present V2CE, a novel pipeline that converts video to high-fidelity event streams for tasks requiring precise events. We introduce quantifiable metrics that elevate this field from qualitative to rigorous scientific analysis. V2CE excels overwhelmingly across all metrics.



\bibliographystyle{IEEEtran}
\bibliography{ref}

\begin{thebibliography}{10}
\providecommand{\url}[1]{#1}
\csname url@rmstyle\endcsname
\providecommand{\newblock}{\relax}
\providecommand{\bibinfo}[2]{#2}
\providecommand\BIBentrySTDinterwordspacing{\spaceskip=0pt\relax}
\providecommand\BIBentryALTinterwordstretchfactor{4}
\providecommand\BIBentryALTinterwordspacing{\spaceskip=\fontdimen2\font plus
\BIBentryALTinterwordstretchfactor\fontdimen3\font minus \fontdimen4\font\relax}
\providecommand\BIBforeignlanguage[2]{{%
\expandafter\ifx\csname l@#1\endcsname\relax
\typeout{** WARNING: IEEEtran.bst: No hyphenation pattern has been}%
\typeout{** loaded for the language `#1'. Using the pattern for}%
\typeout{** the default language instead.}%
\else
\language=\csname l@#1\endcsname
\fi
#2}}

\bibitem{sandamirskaya2022neuromorphic}
Y.~Sandamirskaya, M.~Kaboli, J.~Conradt, and T.~Celikel, ``Neuromorphic computing hardware and neural architectures for robotics,'' \emph{Science Robotics}, vol.~7, no.~67, p. eabl8419, 2022.

\bibitem{zhu2017event}
A.~Z. Zhu, N.~Atanasov, and K.~Daniilidis, ``Event-based feature tracking with probabilistic data association,'' in \emph{2017 IEEE International Conference on Robotics and Automation (ICRA)}.\hskip 1em plus 0.5em minus 0.4em\relax IEEE, 2017, pp. 4465--4470.

\bibitem{li2021tracking}
H.~Li and J.~Stueckler, ``Tracking 6-dof object motion from events and frames,'' in \emph{2021 IEEE International Conference on Robotics and Automation (ICRA)}.\hskip 1em plus 0.5em minus 0.4em\relax IEEE, 2021, pp. 14\,171--14\,177.

\bibitem{gehrig2023recurrent}
M.~Gehrig and D.~Scaramuzza, ``Recurrent vision transformers for object detection with event cameras,'' in \emph{Proceedings of the IEEE/CVF Conference on Computer Vision and Pattern Recognition}, 2023, pp. 13\,884--13\,893.

\bibitem{chamorro2023event}
W.~Chamorro, J.~Sol{\`a}, and J.~Andrade-Cetto, ``Event-imu fusion strategies for faster-than-imu estimation throughput,'' in \emph{Proceedings of the IEEE/CVF Conference on Computer Vision and Pattern Recognition}, 2023, pp. 3975--3982.

\bibitem{Baby_2017}
\BIBentryALTinterwordspacing
S.~A. Baby, B.~Vinod, C.~Chinni, and K.~Mitra, ``Dynamic vision sensors for human activity recognition,'' in \emph{2017 4th {IAPR} Asian Conference on Pattern Recognition ({ACPR})}.\hskip 1em plus 0.5em minus 0.4em\relax {IEEE}, nov 2017. [Online]. Available: \url{https://doi.org/10.1109%2Facpr.2017.136}
\BIBentrySTDinterwordspacing

\bibitem{Gehrig2021DSECAS}
\BIBentryALTinterwordspacing
M.~Gehrig, W.~Aarents, D.~Gehrig, and D.~Scaramuzza, ``Dsec: A stereo event camera dataset for driving scenarios,'' \emph{IEEE Robotics and Automation Letters}, vol.~6, pp. 4947--4954, 2021. [Online]. Available: \url{https://api.semanticscholar.org/CorpusID:232170230}
\BIBentrySTDinterwordspacing

\bibitem{Rebecq2019HighSA}
\BIBentryALTinterwordspacing
H.~Rebecq, R.~Ranftl, V.~Koltun, and D.~Scaramuzza, ``High speed and high dynamic range video with an event camera,'' \emph{IEEE Transactions on Pattern Analysis and Machine Intelligence}, vol.~43, pp. 1964--1980, 2019. [Online]. Available: \url{https://api.semanticscholar.org/CorpusID:189998802}
\BIBentrySTDinterwordspacing

\bibitem{Mahlknecht2022ExploringEC}
\BIBentryALTinterwordspacing
F.~Mahlknecht, D.~Gehrig, J.~Nash, F.~M. Rockenbauer, B.~Morrell, J.~Delaune, and D.~Scaramuzza, ``Exploring event camera-based odometry for planetary robots,'' \emph{IEEE Robotics and Automation Letters}, vol.~7, pp. 8651--8658, 2022. [Online]. Available: \url{https://api.semanticscholar.org/CorpusID:248119129}
\BIBentrySTDinterwordspacing

\bibitem{Seok2020RobustFT}
\BIBentryALTinterwordspacing
H.~Seok and J.~Lim, ``Robust feature tracking in dvs event stream using b{\'e}zier mapping,'' \emph{2020 IEEE Winter Conference on Applications of Computer Vision (WACV)}, pp. 1647--1656, 2020. [Online]. Available: \url{https://api.semanticscholar.org/CorpusID:214676285}
\BIBentrySTDinterwordspacing

\bibitem{Dong2021StandardAE}
\BIBentryALTinterwordspacing
Y.-T. Dong and T.~Zhang, ``Standard and event cameras fusion for feature tracking,'' \emph{Proceedings of the 2021 International Conference on Machine Vision and Applications}, 2021. [Online]. Available: \url{https://api.semanticscholar.org/CorpusID:236435673}
\BIBentrySTDinterwordspacing

\bibitem{Pan2020SingleIO}
\BIBentryALTinterwordspacing
L.~Pan, M.~Liu, and R.~I. Hartley, ``Single image optical flow estimation with an event camera,'' \emph{2020 IEEE/CVF Conference on Computer Vision and Pattern Recognition (CVPR)}, pp. 1669--1678, 2020. [Online]. Available: \url{https://api.semanticscholar.org/CorpusID:214743326}
\BIBentrySTDinterwordspacing

\bibitem{Bardow2016SimultaneousOF}
\BIBentryALTinterwordspacing
P.~Bardow, A.~J. Davison, and S.~Leutenegger, ``Simultaneous optical flow and intensity estimation from an event camera,'' \emph{2016 IEEE Conference on Computer Vision and Pattern Recognition (CVPR)}, pp. 884--892, 2016. [Online]. Available: \url{https://api.semanticscholar.org/CorpusID:10280488}
\BIBentrySTDinterwordspacing

\bibitem{ZHANG2023126388}
\BIBentryALTinterwordspacing
Z.~Zhang, K.~Chai, H.~Yu, R.~Majaj, F.~Walsh, E.~Wang, U.~Mahbub, H.~Siegelmann, D.~Kim, and T.~Rahman, ``Neuromorphic high-frequency 3d dancing pose estimation in dynamic environment,'' \emph{Neurocomputing}, vol. 547, p. 126388, 2023. [Online]. Available: \url{https://www.sciencedirect.com/science/article/pii/S0925231223005118}
\BIBentrySTDinterwordspacing

\bibitem{9959313}
P.~Liu, G.~Chen, Z.~Li, D.~Clarke, Z.~Liu, R.~Zhang, and A.~Knoll, ``Neurodfd: Towards efficient driver face detection with neuromorphic vision sensor,'' in \emph{2022 International Conference on Advanced Robotics and Mechatronics (ICARM)}, 2022, pp. 268--273.

\bibitem{pmlr-v87-rebecq18a}
\BIBentryALTinterwordspacing
H.~Rebecq, D.~Gehrig, and D.~Scaramuzza, ``Esim: an open event camera simulator,'' in \emph{Proceedings of The 2nd Conference on Robot Learning}, ser. Proceedings of Machine Learning Research, A.~Billard, A.~Dragan, J.~Peters, and J.~Morimoto, Eds., vol.~87.\hskip 1em plus 0.5em minus 0.4em\relax PMLR, 29--31 Oct 2018, pp. 969--982. [Online]. Available: \url{https://proceedings.mlr.press/v87/rebecq18a.html}
\BIBentrySTDinterwordspacing

\bibitem{Gehrig_2020_CVPR}
D.~Gehrig, M.~Gehrig, J.~Hidalgo-Carri\'o, and D.~Scaramuzza, ``Video to events: Recycling video datasets for event cameras,'' in \emph{{IEEE} Conf. Comput. Vis. Pattern Recog. (CVPR)}, June 2020.

\bibitem{Hu2021-v2e-cvpr-workshop-eventvision2021}
\BIBentryALTinterwordspacing
Y.~Hu, S.~C. Liu, and T.~Delbruck, ``v2e: From video frames to realistic {DVS} events,'' in \emph{2021 {IEEE/CVF} Conference on Computer Vision and Pattern Recognition Workshops ({CVPRW})}.\hskip 1em plus 0.5em minus 0.4em\relax IEEE, 2021. [Online]. Available: \url{http://arxiv.org/abs/2006.07722}
\BIBentrySTDinterwordspacing

\bibitem{zhu2019eventgan}
A.~Z. Zhu, Z.~Wang, K.~Khant, and K.~Daniilidis, ``Eventgan: Leveraging large scale image datasets for event cameras,'' 2019.

\bibitem{10168206}
Y.~Jiang, Y.~Wang, S.~Li, Y.~Zhang, M.~Zhao, and Y.~Gao, ``Event-based low-illumination image enhancement,'' \emph{IEEE Transactions on Multimedia}, pp. 1--12, 2023.

\bibitem{liu2022lowlight}
L.~Liu, J.~An, J.~Liu, S.~Yuan, X.~Chen, W.~Zhou, H.~Li, Y.~Wang, and Q.~Tian, ``Low-light video enhancement with synthetic event guidance,'' 2022.

\bibitem{Scarselli2009TheGN}
\BIBentryALTinterwordspacing
F.~Scarselli, M.~Gori, A.~C. Tsoi, M.~Hagenbuchner, and G.~Monfardini, ``The graph neural network model,'' \emph{IEEE Transactions on Neural Networks}, vol.~20, pp. 61--80, 2009. [Online]. Available: \url{https://api.semanticscholar.org/CorpusID:206756462}
\BIBentrySTDinterwordspacing

\bibitem{schaefer2022aegnn}
S.~Schaefer, D.~Gehrig, and D.~Scaramuzza, ``Aegnn: Asynchronous event-based graph neural networks,'' 2022.

\bibitem{10166491}
D.~Sun and H.~Ji, ``Event-based object detection using graph neural networks,'' in \emph{2023 IEEE 12th Data Driven Control and Learning Systems Conference (DDCLS)}, 2023, pp. 1895--1900.

\bibitem{slide_gcn}
L.~Yijin, Z.~Han, Y.~Bangbang, C.~Zhaopeng, B.~Hujun, and Z.~Guofeng, ``Graph-based asynchronous event processing for rapid object recognition,'' in \emph{International Conference on Computer Vision (ICCV)}, October 2021.

\bibitem{Tavanaei2018DeepLI}
\BIBentryALTinterwordspacing
A.~Tavanaei, M.~Ghodrati, S.~R. Kheradpisheh, T.~Masquelier, and A.~Maida, ``Deep learning in spiking neural networks,'' \emph{Neural networks : the official journal of the International Neural Network Society}, vol. 111, pp. 47--63, 2018. [Online]. Available: \url{https://api.semanticscholar.org/CorpusID:5039751}
\BIBentrySTDinterwordspacing

\bibitem{Deng2022TemporalET}
\BIBentryALTinterwordspacing
S.-W. Deng, Y.~Li, S.~Zhang, and S.~Gu, ``Temporal efficient training of spiking neural network via gradient re-weighting,'' \emph{ArXiv}, vol. abs/2202.11946, 2022. [Online]. Available: \url{https://api.semanticscholar.org/CorpusID:247084286}
\BIBentrySTDinterwordspacing

\bibitem{cordone2021learning}
L.~Cordone, B.~Miramond, and S.~Ferrante, ``Learning from event cameras with sparse spiking convolutional neural networks,'' 2021.

\bibitem{zhu2022eventbased}
L.~Zhu, X.~Wang, Y.~Chang, J.~Li, T.~Huang, and Y.~Tian, ``Event-based video reconstruction via potential-assisted spiking neural network,'' 2022.

\bibitem{zhu2018multivehicle}
A.~Z. Zhu, D.~Thakur, T.~{\"O}zaslan, B.~Pfrommer, V.~Kumar, and K.~Daniilidis, ``The multivehicle stereo event camera dataset: An event camera dataset for 3d perception,'' \emph{IEEE Robotics and Automation Letters}, vol.~3, no.~3, pp. 2032--2039, 2018.

\end{thebibliography}

\end{document}